\documentclass[letterpaper, 10 pt, conference]{ieeeconf}  

\IEEEoverridecommandlockouts                              

\title{\LARGE \bf
HOGraspFlow: Taxonomy-Aware Hand–Object Retargeting for Multi-Modal SE(3) Grasp Generation 
\thanks{Karlsruhe Institute of Technology, Karlsruhe, Germany. Email: {\tt\small \{name\}.\{surname\}@kit.edu}. \\ This work is supported by the German Federal Ministry of Research, Technology, and Space (BMFTR) under the Robotics Institute Germany (RIG), the DFG SFB-1574-471687386 project, and the Ministry of Science, Research and Arts of the Federal State of Baden-Württemberg within the InnovationCampus Future Mobility.}
}

\author{Yitian Shi, Zicheng Guo, Rosa Wolf, Edgar Welte, Rania Rayyes
}
\usepackage{cuted}
\usepackage{dsfont}
\usepackage{booktabs}
\usepackage{graphicx}
\usepackage{subcaption}
\usepackage{amsmath, amssymb, amsfonts}
\usepackage{mathtools}     
\DeclareMathOperator{\Log}{\text{Log}}
\DeclareMathOperator{\Exp}{\text{Exp}}
\usepackage{graphicx}
\usepackage{cuted}
\usepackage[hidelinks]{hyperref}
\begin{document}
\maketitle
\begin{figure*}
\centering
  \includegraphics[width=.95\textwidth]{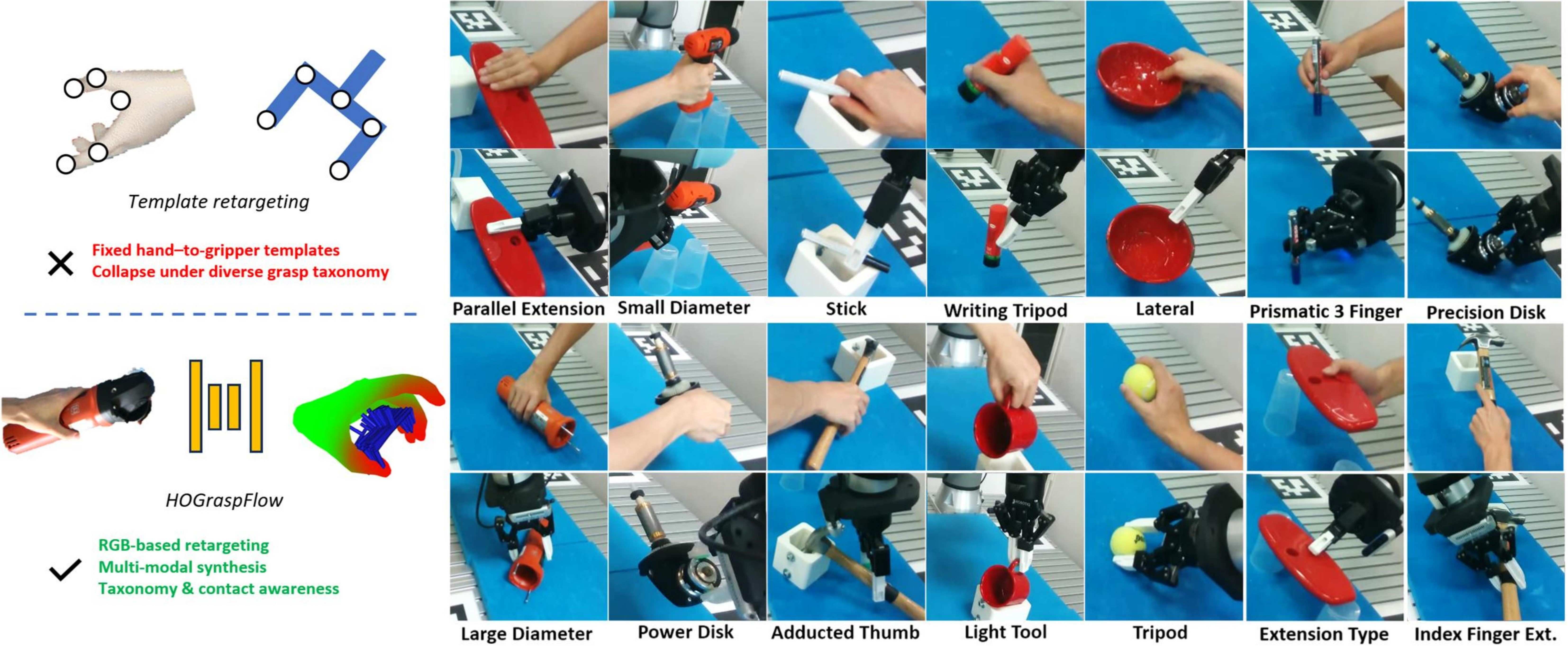}
  \captionof{figure}{Grasp demonstrations for parallel jaw grippers via HOGraspFlow}\label{fig:teaser}
\end{figure*}

\thispagestyle{empty}
\pagestyle{empty}

\bstctlcite{MyBSTcontrol} 
\begin{abstract} We propose Hand-Object\emph{(HO)GraspFlow}, an affordance-centric approach that retargets a single RGB with hand-object interaction (HOI) into multi-modal executable parallel jaw grasps without explicit geometric priors on target objects. Building on foundation models for hand reconstruction and vision, we synthesize $SE(3)$ grasp poses with denoising flow matching (FM), conditioned on the following three complementary cues: RGB foundation features as visual semantics, HOI contact reconstruction, and taxonomy-aware prior on grasp types. Our approach demonstrates high fidelity in grasp synthesis without explicit HOI contact input or object geometry, while maintaining strong contact and taxonomy recognition. Another controlled comparison shows that \emph{HOGraspFlow} consistently outperforms diffusion-based variants (\emph{HOGraspDiff}), achieving high distributional fidelity and more stable optimization in $SE(3)$.
We demonstrate a reliable, object-agnostic grasp synthesis from human demonstrations in real-world experiments, where an average success rate of over $83\%$ is achieved. Code: https://github.com/YitianShi/HOGraspFlow
\end{abstract}

\section{INTRODUCTION}

Recent years have witnessed a growing body of approaches in learning robotic manipulation behaviors directly from human demonstrations, ranging from teleoperation to internet-scale video corpora \cite{11128322,haldar2025point, welte2025interactive}. Specifically, learning from human-object interaction (HOI) demonstrations is increasingly framed as a retargeting problem, where a robotic end–effector (EE) is mapped from anthropometric hand motion discovered in the wild to the EE kinematics. 

Towards the challenge of kinematic mismatch between parallel jaw (PJ) EE and anthropometric hand in the video observations, a popular abstraction in vision-based imitation learning aligns the EE with the human thumb–index pair \cite{11128322, wang2025gat, lepert2025phantom}, enabling simple pinch retargeting. While appealingly concise, this proxy collapses the diversity of human grasp taxonomy \cite{feix2015grasp} and neglects contact–dependent antipodal force closure \cite{bohg2013data}. Hence, this approach is limited to pinch-like grasps and cannot be robustly applied to dynamic demonstrations that exhibit diverse grasp types. 

Moreover, recent research on multi-embodiment grasp generations \cite{khargonkar2023neuralgrasps, attarian2023geometry} has demonstrated that object-conditioned grasp priors can be learned and modulated by geometric embeddings (e.g., Signed Distance Function (SDF), point graph features) that capture EE morphology.
While such approaches enable transfer across different EEs, they are fundamentally object–centric and typically assume access to reliable 3D geometries and pose estimation at test time. Critically, they do not parse human intent or contact semantics from HOI, and thus cannot directly exploit in-the-wild video.

Targeting these limitations, in this work, we develop a vision-based generative hand-to–EE retargeting framework that supports adaptable transfer to dynamic, diverse anthropometric grasp demonstrations, while explicitly recognizing grasp taxonomy and contact with a single RGB crop of the hand. 
Our design is motivated by the asymmetry between intent and object variability, where objects vary widely while the set of underlying grasp intents remains relatively limited. On the object side, geometric diversity (shape, scale, etc.) and sensing brittleness (missing depth, specular failures) induce large, unstructured test-time shifts. In contrast, conditioned on a demonstrated grasp, the space of physically valid human hand poses with contact is constrained by anatomy, kinematics, compact grasp taxonomy, and affordance-driven intent, thus lying on a low-dimensional, structured manifold. 

Moreover, building on recent advances in multi-modal grasp synthesis with diffusion models~\cite{10161569,lim2024equigraspflow,huang2025hgdiffuser}, we adopt denoising generative modeling on $SE(3)$ for intent-conditioned retargeting given the following advantages: (i) Due to the kinematic mismatch between the human hand and PJ grippers, a single human grasp type often corresponds to multiple valid gripper approaches realizing the same affordance. Denoising-based models naturally preserve this multi-modality by sampling diverse modes in $SE(3)$
; (ii) The iterative denoising process admits controllable guidance~\cite{frans2025diffusion}, allowing generated poses to be steered by differentiable constraints (e.g., collision avoidance~\cite{li2025language} or curated affordance priors~\cite{huang2025hgdiffuser}); (iii) In contrast to end-to-end methods that 
rely on post hoc filtering~\cite{sundermeyer2021contact, shi2025vmf}, where feasibility check and affordance matching are performed after grasp generation and invalid candidates are discarded, our in-the-loop guidance enforces constraints during generation, thereby reducing rejections and improving sample efficiency—a critical factor given the scarcity and sampling cost of force-closure, affordance-consistent ground-truth datasets~\cite{10309974}.

In summary, our contributions are:
(i) We designed a vision-based affordance-centric HOI retargeting framework that produces multi-modal 6-DoF PJ grasps from a single RGB frame. This is achieved by conditioning on foundational features on HOI, without requiring explicit object models or pose estimation, while accounting for the diversity of human grasp taxonomy; (ii) We introduce two generative retargeting frameworks, \emph{HOGraspDiff} and \emph{HOGraspFlow}, inspired by SOTA $SE(3)$ generative approaches to a vision-based setting. By integrating state-of-the-art visual foundation models' features as contact priors, our method operates without explicit 3D geometric conditions or pose estimation of objects; (iii) Through extensive ablations and real-world deployment, we demonstrate consistent improvements in grasp-type prediction, contact accuracy, and distributional fidelity relative to the baselines, including a \emph{contact-oracle} variant. With a minor translational correction with depths information at deployment, we achieved over $83\%$ success in grasp transfer in our real-world robot experiments.

\section{Related Works}

\subsection{Generative 6-DoF grasp synthesis}
Learning-based grasp generators model a distribution over executable gripper poses conditioned on scene observations, enabling sampling-based exploration rather than hand-crafted proposal scoring. For instance, GraspNet-1B~\cite{fang2020graspnet} and Contact-GraspNet~\cite{sundermeyer2021contact} have established the effectiveness of learning from local contact geometry for grasp generation. In contrast, denoising-based grasp generators learn a latent density field through an iterative denoising process~\cite{10.3389/frobt.2025.1606247}, representing a recent trend in robotic manipulation learning. Among these, $SE(3)$ diffusion fields~\cite{10161569} adapt denoising diffusion to the Lie group~\cite{sola2018micro}, which learn grasp densities and refine samples directly on the \(SE(3)\) manifold, coupling pose synthesis with motion optimization for grasp execution. To handle symmetries and improve consistency under object rigid motions, EquiGraspFlow~\cite{lim2024equigraspflow} enforces $SE(3)$ equivariance while adopting flow-based denoising to handle symmetries and rigid object motions. To incorporate task constraints and human intent, HGDiffuser~\cite{huang2025hgdiffuser} augments the diffusion process with hand-intent cues, producing task-oriented 6-DoF grasps via guidance~\cite{dhariwal2021diffusion}.
While effective, these generators typically assume accurate object geometry at inference, and thus usually fail under sensing artifacts and large out-of-distribution (OOD) variability in object shape and appearance.
\begin{figure*}[tb]
    \centering
    \includegraphics[width=.94\linewidth]{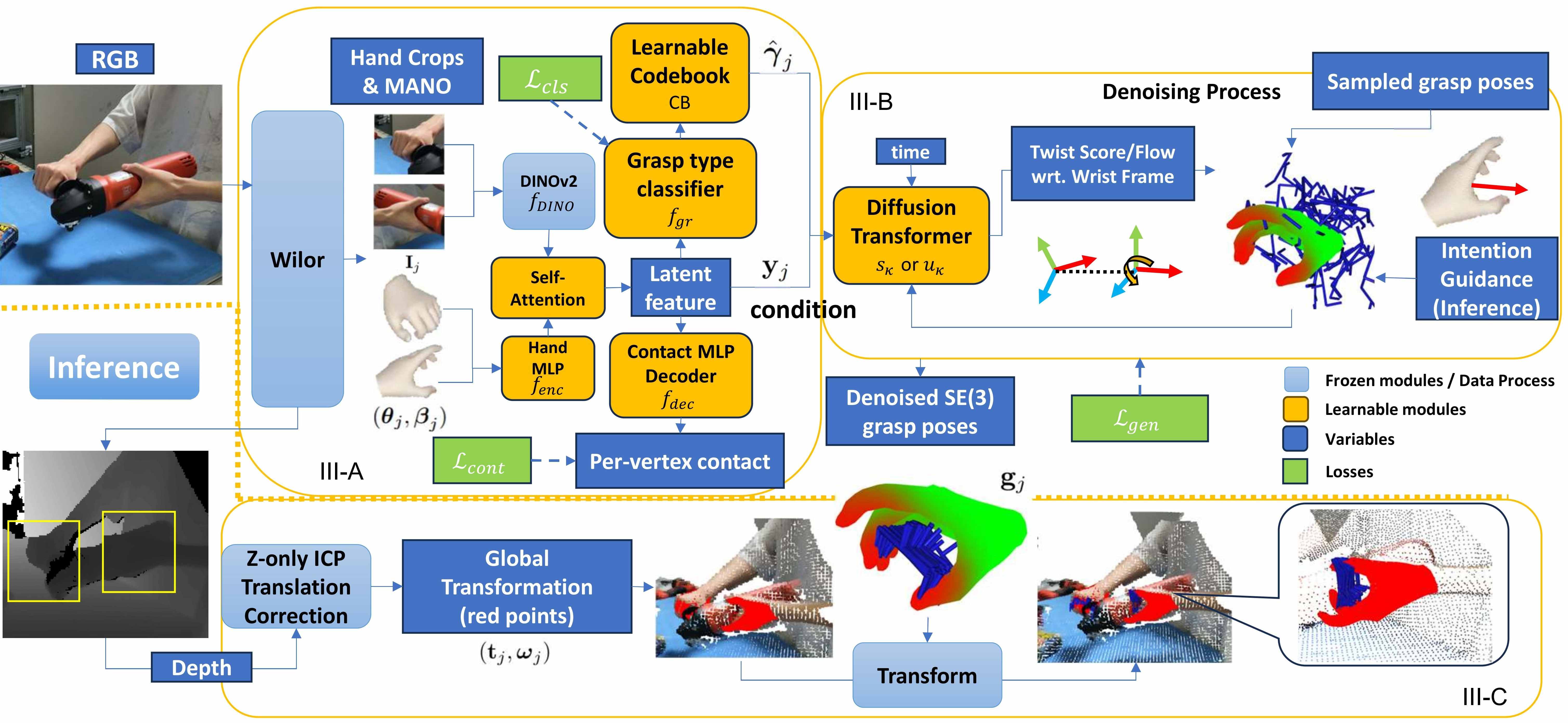}
    \caption{Pipeline for \emph{HOGraspFlow}}
    \label{fig:pipeline}
\end{figure*}
\subsection{Hand-object interaction (HOI) and reconstruction}
Recent progress in HOI recovery has been driven by the MANO parametric hand model~\cite{romero2022embodied} and modern monocular reconstructions that deliver accurate 3D hand pose and shape from RGB~\cite{pavlakos2024reconstructing, potamias2025wilor}. As the foundation, rich HOI datasets provide contact supervision and cross-object variability such as DexYCB~\cite{chao2021dexycb} and OakInk~\cite{yang2022oakink}, which comprises affordances and human interactions over diverse household objects. Meanwhile, HOGraspNet~\cite{cho2024dense} contributes dense HOI annotations with grasp taxonomy in dynamic sequences, facilitating systematic analysis of everyday human grasps. 
Aligned with these trends, we condition grasp generation on reconstructed hand poses and dense contact maps, while deliberately avoiding direct dependence on object geometry, which is prone to reconstruction artifacts and time latency.

\subsection{Learning PJ manipulation from human demonstrations}

Learning from human demonstrations has progressed from teleoperation to in-the-wild human videos, which aim to narrow down the gap between robotic imitation learning and internet-scale demonstrations. As specific instances for learning manipulation with PJ EE, R+x~\cite{11128322} mines large video corpora to retrieve and adapt relevant HOI behaviors, while Point Policy~\cite{haldar2025point} learns visuomotor policies by extracting EE–hand keypoint correspondences from demonstrations. Focusing on grasping, GAT-Grasp~\cite{wang2025gat} conditions on gestures and affordances to translate human hand signals into task-aware robotic grasps.  However, these approaches typically assume a thumb–index pair template as the basis for grasp retargeting, which severely limits their generalizability to diverse and unconstrained in-the-wild demonstrations. 

\section{Methodology}
We aim to recover affordance–centric grasp intent from a human demonstration and retarget it to a PJ, which answers two fundamental questions for affordance-centric grasping: \emph{where to grasp} and \emph{how to grasp}. Fig.~\ref{fig:pipeline} summarizes the full pipeline of our approach. Unlike prior pipelines that require object meshes or partial point clouds at both training and inference~\cite{khargonkar2023neuralgrasps, huang2025hgdiffuser}, we condition grasp generation solely on the outputs of a foundational hand reconstructor WiLoR~\cite{potamias2025wilor}, represented by MANO~\cite{romero2022embodied} parameters and hand detections with semantics from DINOv2~\cite{oquab2023dinov2}. 

To achieve this, our system converts a single RGB observation and the extracted hand poses from WiLoR into a compact grasp–intent embedding (Sec. \ref{subsec:feat}), and then synthesizes multi–modal PJ poses via a generative denoising process in the \(\mathrm{SE}(3)\) manifold (Sec. \ref{subsec:se3}). At inference (Sec. \ref{subsec:deployment}), depth information is only used to refine the absolute translation of the reconstructed hand pose via \emph{Z-only ICP}.

\subsection{Hand-object Perception and Feature Extraction}
\label{subsec:feat}
We first extract HOI semantics from foundational features (Sec.~\ref{parag:feature_ext}).
Then, these representations get refined and guided with two complementary substreams (Sec.~\ref{parag:contact_taxonomy}):
(i) \emph{hand contact estimation}, which principally encodes the localization of feasible grasps by predicting contact maps on the hand surfaces (i.e., answering \emph{where to grasp}); and
(ii) \emph{grasp taxonomy recognition}, which shapes categorical prior on the distribution of PJ grasps, used jointly with a trainable codebook (CB) as a reference embedding (i.e., answering \emph{how to grasp}).

\paragraph{HOI feature extraction}
\label{parag:feature_ext}
Given an RGB image $\mathbf{I}\in\mathbb{R}^{H\times W\times 3}$ with a HOI demonstration,
WiLoR~\cite{potamias2025wilor} detects $n$ hand boxes $\mathcal{B}=\{\mathbf{b}_j\}_{j=1}^{n}$, 
and regresses the global hand translation
$\boldsymbol{t}_j\in\mathbb{R}^{3}$ and MANO parameters
$(\boldsymbol{\omega}_j,\boldsymbol{\theta}_j,\boldsymbol{\beta}_j)$ with global orientation
$\boldsymbol{\omega}_j\in\mathbb{R}^{3}$, joint angles $\boldsymbol{\theta}_j\in\mathbb{R}^{45}$, and shape
$\boldsymbol{\beta}_j\in\mathbb{R}^{10}$. We then crop the image around each detection,
$\mathbf{I}_j=\mathcal{C}(\mathbf{I};\mathbf{b}_j)$, and apply background augmentation using
ground-truth HOI masks from the dataset \cite{cho2024dense}. 

To capture the underlying semantics in HOI, explicit CAD or reconstructed meshes can fail under occlusion, specularity, and category diversity. In contrast, foundation models like DINOv2~\cite{oquab2023dinov2} 
 are broadly invariant across instances and appearances, making them well-suited for grasp representation and transfer without requiring explicit object models or poses. Therefore, we process each augmented crop $\mathbf I_j$ with DINOv2, obtaining patch tokens
$\mathbf{Z}_j=f_{\text{DINO}}(\mathbf{I}_j)\in\mathbb{R}^{P\times D}$ that serve as compact,
semantics-aware descriptors of the local HOI context, where $P$ and $D$ denote the number of patches and the
feature dimensionality, respectively.

We map the hand parameters
$[\boldsymbol{\theta}_j ,\boldsymbol{\beta}_j]\in\mathbb{R}^{55}$ to the same dimension $D$ using a small MLP, yielding a hand feature $\mathbf{h}_j=f_{enc}(\boldsymbol\theta_j, \boldsymbol\beta_j)\in\mathbb{R}^{D}$. Notably, to avoid ambiguity in grasp localization under global hand orientation $\boldsymbol{\omega}$, grasp synthesis is grounded in the hand wrist frame and the feature extractor omits $\boldsymbol{\omega}$, using only articulated pose $\boldsymbol{\theta}$ and shape $\boldsymbol{\beta}$. Thus, the estimated grasp poses can be transformed to the world coordinate via $(\mathbf{t}_j, \boldsymbol\omega_j)$. The feature fusion is performed with a self-attention operator after concatenation between hand and image features\footnote{Only equations referenced in the text are numbered.}:
\begin{equation}
\label{eq:selfattn}
  \mathbf{y}_j \;=\; \operatorname{SelfAttn}\!\big([\mathbf{h}_j,\;\operatorname{ReLU}(f^*(\mathbf{Z}_j))]\big)
\;\in\;\mathbb{R}^{D},  
\end{equation}
the output $\mathbf{y}_j$ is a single HOI-aware descriptor and $f^*$ is the feature adaptation layers. 

\paragraph{Hand contact and grasp taxonomy recognition}
\label{parag:contact_taxonomy}
To enrich the hand representation, we learn a contact-aware HOI embedding with a lightweight MLP
decoder adapted from Prakash et al.~\cite{prakash2025synthesizing}, without requiring explicit contact input.

Principally, MANO parametric model \cite{romero2022embodied} map the hand pose parameters $(\boldsymbol{\theta},\boldsymbol{\beta})$ to compact hand meshes via: $M = \mathcal{M}(\boldsymbol\theta, \boldsymbol\beta)\in \mathbb{R}^{N_v \times 3}$, with $N_v = 778$ vertices in structured order.
Given the fused descriptor $\mathbf{y}_j$, the decoder 
$f_{\mathrm{dec}}:\mathbb{R}^{D}\!\to\!\mathbb{R}^{N_v}$ predicts per-vertex contact logits $\hat{\mathbf{c}}_j$ over $M$:
\[
\hat{\mathbf{c}}_j \;=\; \operatorname{Sigmoid}\!\big(f_{\mathrm{dec}}(\mathbf{y}_j)\big)\in[0,1]^{N_v}.
\]
Given the per-vertex contact labels $\mathbf{c}_j\in[0,1]^{N_v}$ that are retrieved following \cite{cho2024dense}, we optimize a weighted binary cross entropy:
\[
\mathcal{L}_{\mathrm{cont}}
= -\frac{1}{N_v}\sum_{v=1}^{N_v}
\Big[\, w_1\,\mathbf{c}_{jv}\log \hat{\mathbf{c}}_{jv} +\,(1-\mathbf{c}_{jv})\log (1-\hat{\mathbf{c}}_{jv}) \Big],
\]
$w_1=5$ is chosen to address class imbalance empirically.

While the \emph{hand contact estimation} captures instance-specific HOI semantics over contact localizations, grasp retargeting between the anthropometric hand and PJ is underconstrained due to their morphological differences and kinematic mismatch. To mitigate this, we leverage the \emph{grasp taxonomy recognition} as the morphological prior to complement the retargeted PJ grasp distributions. 

Specifically, a grasp type classifier MLP: $f_{\mathrm{gr}}(\mathbf{y}_j)$ categorizes $K=33$ grasp types, defined by the GRASP Taxonomy~\cite{feix2015grasp}, and trained via cross entropy: $\mathcal{L}_\text{cls} = \text{CE}(f_{\mathrm{gr}}(\mathbf{y}_j), \,cls_j)$ with ground-truths class labels $cls_j$. We maintain a learnable codebook $\text{CB}=\{\boldsymbol{\boldsymbol \gamma}_k\in\mathbb{R}^{D}\}_{k=1}^{K}$ of size $K$, and obtain a semantics-aware prior as a softmax-weighted mixture to mitigate the classification errors:
\[
\boldsymbol{\pi}_j=\operatorname{Softmax}\!\big(f_{\mathrm{gr}}(\mathbf{y}_j)\big),
\qquad
\hat{\boldsymbol{\boldsymbol\gamma}}_j=\sum_{k=1}^{K}\boldsymbol\pi_{j,k}\,\boldsymbol{\boldsymbol\gamma}_k.\]
This taxonomy-conditioned prior complements the contact embedding by regularizing both contact topological and grip orientational distributions.

\subsection{$SE(3)$ Pose Synthesis via Generative Denoising}
\label{subsec:se3}
Given the HOI-aware descriptor $\mathbf{y}_j$ and the induced CB embedding $\hat{\boldsymbol\gamma}_j$ from Sec.~\ref{subsec:feat}, we aim to generate diverse, retargeted $SE(3)$ grasp poses. We deliberately adopt denoising-based generative models since they preserve multi-modality by sampling diverse modes in $SE(3)$. Moreover, iterative samplers admit controllable, differentiable guidance~\cite{frans2025diffusion}, enforcing feasibility during generation and improving sample efficiency over post hoc filtering~\cite{sundermeyer2021contact}.

We introduce two frameworks  \emph{HOGraspDiff} and \emph{HOGraspFlow} that follow the leading families of recent denoising generative models: score-based diffusion~\cite{NEURIPS2019_3001ef25, 10161569} and flow matching~\cite{lipman2022flow}  in $SE(3)$, respectively, both based on the Diffusion Transformer (DiT) 
architecture~\cite{peebles2023scalable}.

We formulate the hand grasp retargeting as a conditional sampling process from a $SE(3)$ pose distribution of PJ:
\[\mathbf{g}_j := (\mathbf{p}_j, \mathbf{q}_j) 
    \sim p(\mathbf{g} \mid \mathbf{y}_i,\, \hat{\boldsymbol\gamma}_j), 
    \quad \mathbf{g} \in SE(3),\]
where $\mathbf{p}_j \in \mathbb{R}^3$ denotes the Euclidean position, and 
$\mathbf{q}_j \in S^3 \subset \mathbb{R}^4$ is the unit quaternion of orientation. Since the reference frame of the denoising process is constructed with respect to the hand wrist frame, the framework naturally inherits equivariance to the global transformation $(\mathbf{t}_j,\boldsymbol{\omega}_j)$. 

\paragraph{HOGraspDiff with score matching (SM)}
We model the forward diffusion as left-invariant Brownian motion in score matching (SM), by pushing forward isotropic Wiener noise on the Lie algebra \(\mathfrak{se}(3)\) analogous to~\cite{ryu2024diffusion}. 

Specifically, let \(\mathbf{g}_t \in SE(3)\) denote the grasp pose at denoising time \(t \in [0,1]\) in the forward diffusion process. We learn a time-dependent left-invariant score function \footnote{We omit notations $\mathbf y_j$, $\hat{\gamma_j}$ as conditions in $s_\kappa(\cdot)$, $u_\kappa(\cdot)$ for simplicity} \(s_{\kappa}(\mathbf{g}_t,t)\approx \nabla_{\mathbf{g}_t}\log p_t(\mathbf{g}_t)\in\mathfrak{se}(3)\) parametrized by $\kappa$. The learned score then drives the reverse-time diffusion:
\begin{equation}
\mathrm{d}\mathbf{x}_t
=u_{\kappa}(\mathbf{g}_t,t)\,\mathrm{d}t
\;+\;
\sqrt{2\,\beta(t)}\cdot\mathrm{d} W_t ,
\label{eq:reverse-sde-se3}
\end{equation}
where \(\beta(t)\ge 0\) is the diffusion schedule and \( W_t\) is left-invariant Wiener process on \(\mathfrak{se}(3)\).

\begin{figure}[t]
    \centering        \includegraphics[width=1.\linewidth]{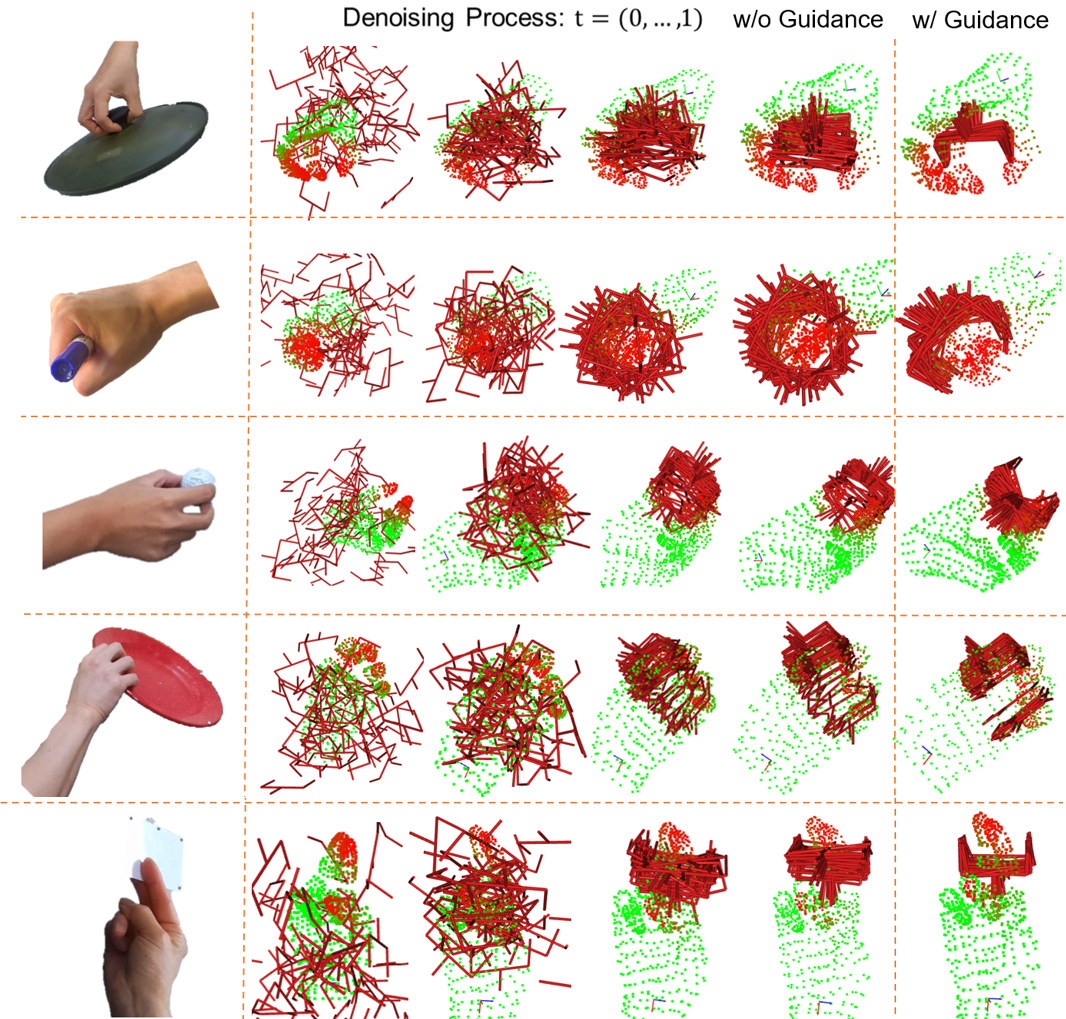}
        \caption{Denoising process and generation results. Vertices in contact are in red. Parameters for guidance: $\theta_{\rm thr} = 0.8$, $\lambda^{gd}=1e-3$.}
        \label{fig:qual1}
\end{figure}
In training, a displacement 
$\Delta \mathbf{g}_t = (\Delta \mathbf{p}_t, \Delta \mathbf{q}_t) \in SE(3)$ 
is sampled at a random noise level $t\sim\mathcal{U}(0, 1]$, with decomposed translational and rotational elements \cite{ryu2024diffusion}:
\begin{equation}
    p_t(\Delta\mathbf{p}_t) = \mathcal{N}(\mathbf{p}; \mathbf{0}, 
    \sigma_t^2 \mathbf{I}), \,
    p_t(\Delta\mathbf{q}_t) = \mathcal{IG}_{SO(3)}(R_\mathbf{q}; \epsilon_t).
    \label{eq:iso_dist}
\end{equation}
Here, $\mathcal{N}$ and $\mathcal{IG}_{SO(3)}$ denote the isotropic Gaussian distributions in $\mathbb{R}^3$ and $SO(3)$, parameterized by their respective concentration $\sigma_t^2 = \alpha_\mathbf{p}t$ and $\epsilon_t = \tfrac{\alpha_\mathbf{q} t}{2}$. \(R_\mathbf{q}\in SO(3)\) is the rotation matrix of \(\mathbf{q}\). The grasp poses of each time step $\mathbf{g}_t$ is then diffused by the twist
in body frame using group product\footnote{$\Log:SO(3)\!\to\!\mathfrak{so}(3)$ and $\Exp:\mathfrak{so}(3)\!\to\!SO(3)$ denote the logarithm and exponential maps, whereas "$\log$" is the standard scalar logarithm.}:
\begin{equation}
\mathbf{g}_{t+\Delta t} 
= \mathbf{g}_t \circ \Delta\mathbf{g}_{t}
=:
\begin{bmatrix}
\mathbf{p}_t + R_{\mathbf{q}_t}\,J(\Log(R_{\Delta \mathbf{q}_t}))\,\Delta \mathbf{p}_t \\[6pt]
\mathbf{q}_t \otimes \Delta \mathbf{q}_t
\end{bmatrix},
\label{eq:exp}
\end{equation} 
where \(\otimes\) denotes quaternion multiplication and $J(\cdot)$ is the left Jacobian. Hence, the objective is to learn a score head that estimates the translational and
rotational scores, namely $s^\mathbf{p}_{\kappa}(\mathbf{g}_t, t)$ and $s^\mathbf{q}_{\kappa}(\mathbf{g}_t, t)$, 
by minimizing the mean squared error to the ground-truth scores:
\begin{align*}
\mathcal{L}_{score}
= \mathbb{E}_{\mathbf{g}_0,\,t,\,\Delta \mathbf{g}_t}\!\Big[
  &\big\|\,\nabla_{\Delta \mathbf{p}_t}\log p_t(\Delta\mathbf{p}_t)
      - s^\mathbf{p}_{\kappa}(\mathbf{g}_t,t)\,\big\|_2^2 \\
  +\;&\big\|\,\nabla^{\mathbb{L}}_{\Delta \mathbf{q}_t}\log p_t(\Delta\mathbf{q}_t)
      - s^\mathbf{q}_{\kappa}(\mathbf{g}_t,t)\,\big\|_2^2
\Big].
\end{align*}
Closed-form solutions are calculated via \cite{ryu2024diffusion}:
\begin{align*}
\nabla_{\Delta\mathbf{p}_t}\log p_t(\Delta\mathbf{p}_t)
&= -\Delta \mathbf{p}_t\sigma^{-2}_t,
\\[4pt]
\nabla^{\mathbb{L}}_{\Delta\mathbf{q}_t}\log p_t(\Delta\mathbf{q}_t)
&= \sum_{i=1}^{3}
\mathbb{L}_i\,\log\mathcal{IG}_{SO(3)}\!\big(R_{\Delta \mathbf{q}_t};\,\epsilon_t\big)\,\mathbf{e}_i,
\end{align*}
$\mathbb{L}_i$ is the left-trivialized Lie derivative of~$\mathcal{IG}_{SO(3)}(R_\mathbf{q}; \epsilon)$ along the $i$-th orthogonal basis \(\{\mathbf e_i\}_{i=1,2,3}\) on \(\mathfrak{so}(3)\).

In sampling, for each translational or rotational element $\mathbf{x}\in\{\mathbf{p},\mathbf{q}\}$ , the update increment follows Eq.~\eqref{eq:reverse-sde-se3}:
\begin{align*}
    \Delta \mathbf{x}_t
    \;=\; \beta_{\mathbf{x}}(t)\,s_{\kappa}^{\mathbf{x}}(\mathbf{g}_t,t)\,\Delta t
    \;+\sqrt{2\beta_{\mathbf{x}}(t)}\mathbf{z}_{\mathbf{x},t},
    \label{eq:sampling_score}
\end{align*}
where $\beta_{\mathbf{x}}(t)=\frac{1}{2}\alpha^2_{\mathbf{x}}\,t^{\alpha_t}$ with $\alpha_t$ as the time exponent and the sampling stochasticity of Wiener process: $\mathbf{z}_{\mathbf{x},t}\sim\mathcal{N}(\mathbf{0}, \mathbf{I}^3)$.
We then update the pose by left multiplication iteratively via:
\(
\mathbf{g}_{t-\Delta t} \;=\; \mathbf{g}_{t} \circ \Delta \mathbf{g}_t^{-1}
\) with $SE(3)$ product in Eq.~\eqref{eq:exp}.

\paragraph{HOGraspFlow with flow matching (FM)}
We adapted the SM's formulations to construct the flow-based alternative, which learns a deterministic flow and transports the mass from a base distribution $P_0$ to $P_1 = P_{\text{data}}$. 

Principally, flow matching (FM) parametrizes the drift $\beta_{\mathbf{x}}(t)\,s_{\kappa}^{\mathbf{x}}(\mathbf{g}_t,t)$ by learning left-trivialized velocity components \(u^{\mathbf{p}}_{\kappa}(\mathbf g_t,t)\) and \(u^{\mathbf{q}}_{\kappa}(\mathbf g_t,t)\) along the geodesics in $\mathfrak{se}(3)$, bypassing noise schedules and score normalization.

The smooth time-dependent linear and angular velocity fields~$u^{\mathbf{p}}_{\kappa}(\mathbf{g}_t,t), u^{\mathbf{q}}_{\kappa}(\mathbf{g}_t,t): [0, 1] \times SE(3) \rightarrow \mathbb{R}^3$, mapping from~$\mathbf{g}_t, t\in [0, 1]$ to~$\mathbf g_1 \sim P_1$ are learned, such that poses~$\mathbf g_1$ follow the distribution of the ground-truth data. Therefore, the deterministic flow is constructed and applied via:
\begin{equation}
  \mathbf g_t=
\mathbf g_0 \circ [u^{\mathbf{p}}_{\kappa}(\mathbf{g}_t,t),\, u^{\mathbf{q}}_{\kappa}(\mathbf{g}_t,t)]^\top.
\label{eq:flow-ode}  
\end{equation}
While the body twist naturally unifies translation and rotation, the fully coupled formulation in Eq.~\eqref{eq:flow-ode} creates strong correlations between them, often impeding stable optimization and denoising process. We therefore adopt a decoupled product manifold flow on
$\mathbb R^3\times SO(3)$ with independent priors $\mathbf p_0$ and $\mathbf q_0$ that are sampled same as Eq.~\eqref{eq:iso_dist}. Given a ground-truth pose $\mathbf g_1=( \mathbf{p}_1,\mathbf q_1)$, the geodesics can then be calculated as (in body frame):
\begin{align}
\Delta\mathbf p = R_{\mathbf q_0}^\top(\mathbf {p}_1-\mathbf{p}_0), \,
\Delta\boldsymbol\phi \;=\; \Log(R_{\mathbf q^{-1}_0\mathbf q_1}).
\label{eq:decoupled_flow}
\end{align}
Subsequently, the interpolation between the initial grasp pose~$\mathbf g_0$ and the ground-truth grasp pose~$\mathbf g_1 \sim P_1$ is performed to get the linear and angular transformation approaching the ground-truth value:
\begin{align}
\mathbf{p}_{t+\Delta t} = \mathbf{p}_t +\Delta tR_{\mathbf q_0}\Delta\mathbf{p}, \,
\mathbf q_{t+\Delta t} = \mathbf{q}_t \otimes \Exp(\Delta t \Delta\boldsymbol{\phi}).
\label{eq:flow_sample}
\end{align}
In this way, the translational field is independent of the initial rotation $\mathbf q_0$. The objective is to minimize the mean squared error between the ground-truth velocities and the predictions:
\[
    \mathcal{L}_{flow}
= \mathbb{E}_{\mathbf{g}_0,\,t,\,\Delta \mathbf{g}_t}\!\Big[ | |  \Delta\mathbf{p} - u^{\mathbf{p}}_{\kappa}(\mathbf{g}_t,t)| |^2_2 + | | \Delta\boldsymbol{\phi} - u^{\mathbf{q}}_{\kappa}(\mathbf{g}_t,t)| |^2_2\Big].
\]

At inference, a grasp pose is calculated by sampling an initial pose and solving an ordinary differential equation (ODE) over~$t \in [0, 1]$. We use either 4th-order Runge–Kutta \cite{NIPS2014_fa042d3a} or the generic Euler sampling as Eq.~\eqref{eq:flow_sample} iteratively over $t$, trading off between sample quality and efficiency.

\paragraph{Guidance-based sampling}
To better align the synthesized PJ grasps with the hand intention, guidance is applied in both SM and FM. In the hand frame, $\mathbf e_{\text{app}}=[0, 1, 0]^\top$ is empirically taken as the palm's axis in the local wrist frame. For a noisy pose $\mathbf g_t$, we apply soft guidance only when angular alignment (in cosine similarity) exceeds a threshold:
\[\boldsymbol\xi_t=\nabla^{\mathbb L}_{\mathbf q_t} c_t  \mathds 1[c_t<\theta_{\rm thr}],\quad c_t = \langle \mathbf R_{\mathbf q_t}[y],\mathbf e_{\text{app}}\rangle.
\]
With guidance weight $\lambda^\text{gd}>0$, during sampling the rotational score/flow field is superposed via:
\[
s_{\kappa}^{\mathbf q}(\mathbf g_t,t)
\leftarrow s_{\kappa}^{\mathbf q}(\mathbf g_t,t) + \lambda^{\mathrm{gd}}\,\boldsymbol\xi_t,\,
u_{\kappa}^{\mathbf q}(\mathbf g_t,t)
\leftarrow u_{\kappa}^{\mathbf q}(\mathbf g_t,t) + \lambda^{\mathrm{gd}}\,\boldsymbol\xi_t.
\]
We additionally apply classifier-free guidance \cite{ho2022classifier} by sampling a weighted sum of the conditional and unconditional flow inspired by \cite{lim2024equigraspflow} for both approaches.
The generation processes are illustrated in Fig.~\ref{fig:qual1} for \emph{HOGraspFlow}.
\subsection{Deployment}
Given grasp candidates $\mathbf{g}_j$ sampled in the wrist frame, we transform them to the world frame via the hand's world pose $(\mathbf{t}_j, \boldsymbol{\omega}_j)$. In particular, to compensate for the translation bias of WiLoR in $\mathbf{t}_j$~\cite{wang2025gat, lepert2025phantom}, we apply and constrain the \emph{ICP} algorithm~\cite{segal2009generalized} that only allows correction along the ray from the camera origin to the hand center (\emph{“Z-only ICP”}), which enables accurate 3D hand pose with the monocular recognition, in contrast to multi-view setups in \cite{haldar2025point, wang2025gat}. In addition, this retargeting strategy improves generalization by decoupling grasp generation from the object’s absolute pose. 

\label{subsec:deployment}

\section{Experiments}
In the experiments, we aim to evaluate our framework along the following factors regarding the framework design: (i) the advantage of foundation vision features for synthesized grasp quality and representation; (ii) the relative performance of \emph{HOGraspFlow} versus \emph{HOGraspDiff} under matched settings; (iii) the real-world performance and generalization to unseen objects. We report distributional fidelity of synthesized grasps, per-vertex contact errors, and grasp-type classification scores in Sec.~\ref{subsec:performance1}. To further assess generalization, we perform real-world evaluations to examine our approach on various objects and grasp types in Sec.~\ref{subsec:performance2}.

\subsection{Data preparation}
\begin{figure}
    \centering
    \includegraphics[width=1.\linewidth]{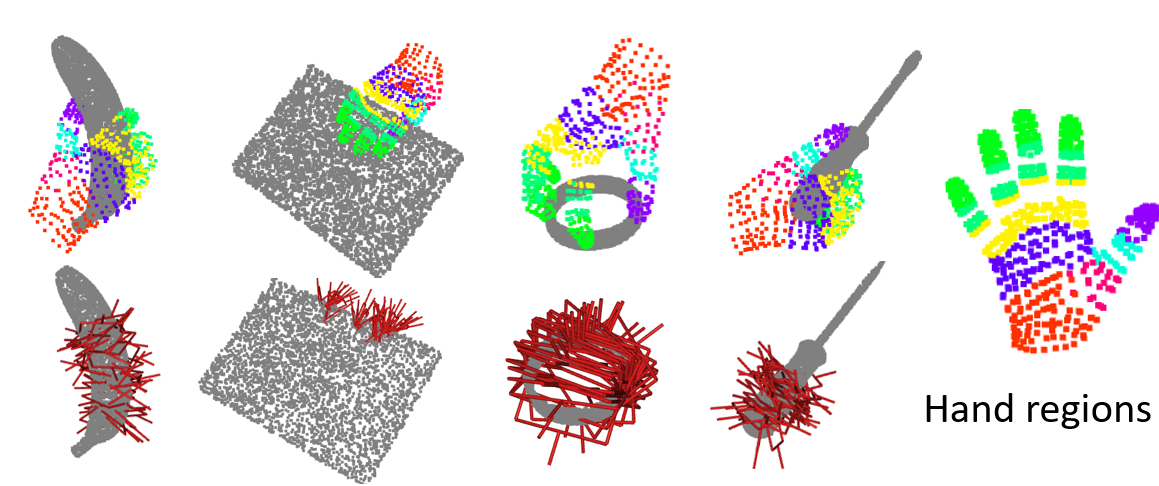}
    \caption{Generated PJ grasps via \emph{region–conditioned contact matching} with HOGraspNet annotations~\cite{cho2024dense} via hand regions defined by~\cite{yang2021cpf}.}
    \label{fig:pj_regions}
\end{figure}

We evaluate our method on the HOGraspNet dataset~\cite{cho2024dense}, which provides $1.5$M HOI demonstrations in RGB-D frames, annotated with MANO hand parameters, object meshes and poses, contact maps, and grasp labels defined by GRASP taxonomy~\cite{feix2015grasp}. To generate supervision of PJ grasp distributions systematically, we synthesize grasp annotations offline in the following two steps: First, PJ grasps are sampled using the MetaGraspNet workflow~\cite{10309974} on each object mesh from the HOGraspNet. This produces collision-free grasps that fulfill antipodal constraints \cite{bohg2013data} and are agnostic to the affordance. Second, given the ground-truth hand vertices and their poses relative to the objects, the sampled PJ grasps are filtered via \emph{region–conditioned contact matching} (Fig.~\ref{fig:pj_regions}): Specifically, we adopt the A-MANO~\cite{yang2021cpf} semantic partition of the MANO surface into $16$ disjoint regions (Fig.~\ref{fig:pj_regions}, right) as a prior for filtering. A pre-generated force-closure candidate is accepted only if its implied contacts can be assigned to two distinct hand regions, reflecting the hypothesis that a realizable PJ grasp engages at least 2 opposing anatomical areas rather than a single region. 


\begin{table}[h!]
\setlength{\tabcolsep}{7pt}
\centering
\begin{tabular}{lccc}
\toprule
Baseline  &  TA(\%)$\uparrow$   &  CA(\%)$\uparrow$   &  EMD$\downarrow$  \\
\midrule
\emph{HOGraspDiff}, w/o DINOv2   & $69.4 $ & $  71.8$ & $1.02 \pm0.22$ \\
w/ DINOv2  & $ 90.3   $ & $ 88.8  $ &$ 0.82\pm 0.23$ \\
w/ DINOv2 CB  & $\mathbf{91.4}   $ & $ 89.5  $ &  $\mathbf{0.80\pm0.23}$ \\
\midrule
w/ Contact  & $78.8  $ & $ \mathbf{90.5}  $ & $0.88\pm0.22$ \\
\midrule
\midrule
\emph{HOGraspFlow}, w/o DINOv2  & $ 68.9   $ & $ 69.2   $ &  $0.81 \pm  0.24$\\
w/ DINOv2  & $ 91.4   $ & $ 87.6  $ & $ 0.69 \pm 0.21  $\\
w/ DINOv2 CB  & $\mathbf{ 92.5   }$ & $ 88.1 $ &$ \mathbf{0.67\pm 0.21} $ \\
\midrule
w/ Contact  & $ 78.5   $ & $ \mathbf{90.5}   $ &$ 0.76 \pm 0.22 $ \\
\midrule
\bottomrule
\end{tabular}
\caption{Comparison of Taxonomy Accuracy (TA(\%))), Contact Accuracy (CA(\%)), and earth mover's distance (EMD).}
\label{tab:overall}
\end{table}

\begin{table*}[h!]
\centering
\setlength{\tabcolsep}{.1pt}
\begin{tabular}{lcccccccccccccccc}
\toprule
   &  \multicolumn{2}{c}{\emph{Small Diameter}}  & \multicolumn{2}{c}{\emph{Parallel Extension}}  & \multicolumn{2}{c}{\emph{Medium Wrap}}   &  \multicolumn{2}{c}{\emph{Sphere 4 Finger}}  & \multicolumn{2}{c}{\emph{Lateral}}   &  \multicolumn{2}{c}{\emph{Stick}}   & \multicolumn{2}{c}{\emph{Palmar Pinch}} &  \multicolumn{2}{c}{\emph{Tripod}} \\
\cmidrule(lr){2-3}\cmidrule(lr){4-5}\cmidrule(lr){6-7}\cmidrule(lr){8-9}\cmidrule(lr){10-11}\cmidrule(lr){12-13}\cmidrule(lr){14-15}\cmidrule(lr){16-17}
  &  CA(\%)$\uparrow$  &  EMD$\downarrow$  &  CA(\%)$\uparrow$ &  EMD$\downarrow$ &  CA(\%)$\uparrow$  &  EMD$\downarrow$  &  CA(\%)$\uparrow$  &  EMD$\downarrow$  &  CA(\%)$\uparrow$  &  EMD$\downarrow$   &  CA(\%)$\uparrow$  &  EMD$\downarrow$  &  CA(\%)$\uparrow$  &  EMD$\downarrow$ &  CA(\%)$\uparrow$  &  EMD$\downarrow$ \\

\midrule
\emph{HOGraspDiff}, w/o DINOv2  & $  82.1$ & $  0.93 $ & $ 84.9 $ & $  1.24 $ & $ 77.7  $ & $ 0.88$ & $ 84.1$ & $ 0.86 $ & $ 60.9 $ & $  1.22 $ & $ 74.0  $ & $0.94 $ & $ 69.4 $ & $ 0.91$ & $  71.6 $ & $ 0.92 $\\

w/ DINOv2   & $ 88.1  $ & $ 0.85  $ & $ 88.9  $ & $ 0.71  $ & $ 85.3  $ & $ 0.87  $ & $ 88.2  $ & $ 0.75  $ & $ 89.5  $ & $ \mathbf{0.72}  $ & $ 84.1  $ & $ 0.86  $ & $ 91.3  $ & $ 0.85  $ & $ 84.8  $ & $ 0.80 $\\
w/ DINOv2 CB   & $ \mathbf{90.6}  $ & $ \mathbf{0.71}  $ & $ 88.5  $ & $ \mathbf{0.67}  $ & $ 87.0  $ & $ \mathbf{0.86}  $ & $ 88.2  $ & $ \mathbf{0.70}  $ & $ 89.3 $ & $ \mathbf{0.72}  $ & $ 87.2  $ & $ \mathbf{0.83}  $ & $ 86.7  $ & $ \mathbf{0.84}  $ & $ 87.6  $ & $ \mathbf{0.77} $\\
\midrule
w/ Contact   & $ 90.1  $ & $ 0.79  $ & $ \mathbf{90.1}  $ & $ 0.79  $ & $ \mathbf{90.4}  $ & $ 0.88  $ & $ \mathbf{92.2}  $ & $ 0.71  $ & $ \mathbf{89.9}  $ & $ 0.78  $ & $ \mathbf{89.3}  $ & $ 0.92  $ & $ \mathbf{95.9}  $ & $ 0.89  $ & $ \mathbf{91.5}  $ & $ 0.84 $\\
\midrule
\midrule
\emph{HOGraspFlow}, w/o DINOv2  & $ 84.5  $ & $ 0.75  $ & $ 86.9  $ & $ 0.82  $ & $ 78.9$   & $ 0.75  $ & $ 84.6  $ & $ 0.69  $ & $ 72.2  $ & $ 0.77  $ & $ 69.6  $ & $ 0.80  $ & $ 88.5  $ & $ 0.73  $ & $ 83.0  $ & $ 0.75 $\\

w/ DINOv2   & $ 87.2  $ & $ 0.63  $ & $ 86.5  $ & $ 0.66  $ & $ 84.6  $ & $ 0.63  $ & $ 84.8  $ & $ \mathbf{0.66}  $ & $ 88.6  $ & $ 0.69  $ & $ 87.5  $ & $ 0.69  $ & $ 92.5  $ & $ 0.70  $ & $ 88.9  $ & $ 0.67 $\\
w/ DINOv2 CB   & $ 87.0  $ & $\mathbf{0.61}$ & $ 88.1  $ & $ \mathbf{0.64}  $ & $ 84.6  $ & $ \mathbf{0.62}  $ & $ 87  $ & $ 0.67  $ & $ 89.6  $ & $ \mathbf{0.67}  $ & $ 86.1  $ & $ \mathbf{0.66}  $ & $ \mathbf{95.0}  $ & $ \mathbf{0.67}  $ & $ 89.9  $ & $ \mathbf{0.61} $\\
\midrule
w/ Contact   & $ \mathbf{89.1}  $ & $ 0.66  $ & $ \mathbf{91.7}  $ & $ 0.71  $ & $ \mathbf{88.0}  $ & $ 0.66  $ & $ \mathbf{91.1}  $ & $ 0.67  $ & $ \mathbf{89.9}  $ & $ 0.72  $ & $ \mathbf{90.1}  $ & $ 0.74  $ & $ 91.7  $ & $ 0.71  $ & $ \mathbf{91.8}  $ & $ 0.73 $\\
\midrule
\bottomrule
\end{tabular}
\caption{8 typical grasp types and their corresponding contact accuracy and EMD.}
\label{tab:per_type}
\end{table*}
\begin{figure*}[t]
    \centering
    \hspace{-.8em}
    \begin{subfigure}{0.50\linewidth}
        \centering
        \includegraphics[width=\linewidth]{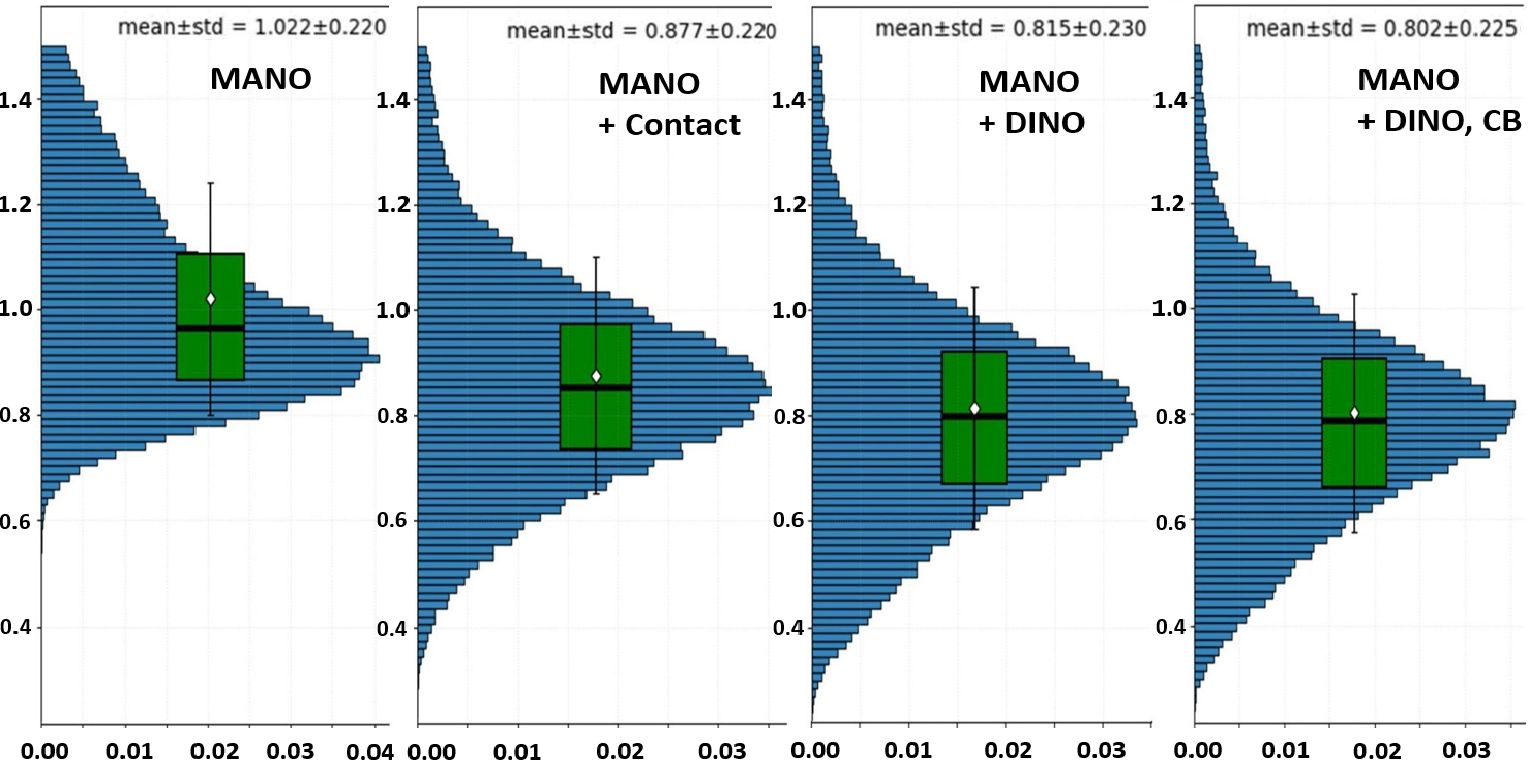}
        \caption{EMD for \emph{HOGraspDiff}}
        \label{fig:flow_emd}
    \end{subfigure}
    \hfill
    \hspace{-1.2em}
    \begin{subfigure}{0.485\linewidth}
        \centering
        \includegraphics[width=\linewidth]{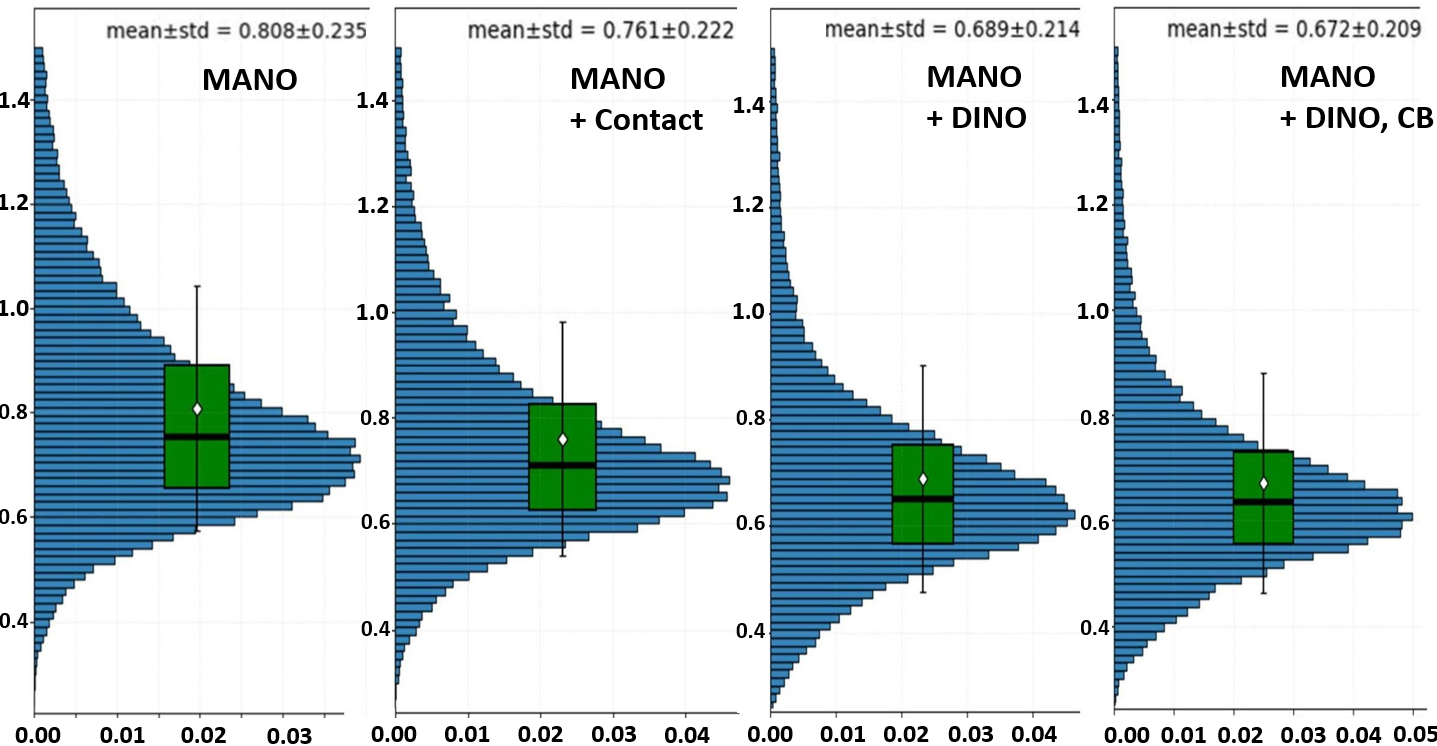}
        \caption{EMD for \emph{HOGraspFlow}}
        \label{fig:other_emd}
    \end{subfigure}
    \caption{Comparison of EMD histograms between \emph{HOGraspDiff} and \emph{HOGraspFlow} on HOGraspNet (in frequency).}
    \label{fig:emd_dist}
\end{figure*}
\subsection{Grasp generation performance with ablations}
\label{subsec:performance1}
\paragraph{Objective}
We train perception and generation end–to–end with \(\mathcal{L}=\lambda_{\mathrm{cls}}\mathcal{L}_{\mathrm{cls}}+\lambda_{\mathrm{cont}}\mathcal{L}_{\mathrm{cont}}+\lambda_{\mathrm{gen}}\mathcal{L}_{\mathrm{gen}}\)
Depending on the chosen generator, either the score or flow-based objective is enabled \(\mathcal{L}_{\mathrm{gen}}\in\{\mathcal{L}_{\mathrm{flow}},\;\mathcal{L}_{\mathrm{score}}\}\). We choose \(\lambda_{\mathrm{cls}} =0.1\), \(\lambda_{\mathrm{cont}}=0.1\) and \(\mathcal{L}_{\mathrm{gen}}=1\) in all baselines.
\paragraph{Baselines}
Our first study probes two design choices in the framework: 2D vision features and the taxonomy-aware codebook. For this purpose, we evaluate four variants both on \emph{HOGraspDiff} and \emph{HOGraspFlow}: (i) \emph{w/o DINOv2}: remove DINO-based self-attention and condition only on the hand MLP feature (i.e. $\mathbf y_j=\mathbf h_j$ in Eq.~\eqref{eq:selfattn}), while holding identical architecture; (ii) \emph{w/ DINOv2}: restore self-attention to DINOv2 patch tokens on the RGB hand crops; (iii) \emph{w/ DINOv2 CB}: further augment with the taxonomy-aware trainable codebook (CB); and iv) \emph{w/ Contact} (or \emph{contact-oracle}): replace visual conditioning with the ground-truth per-vertex contact signal (with $N_v$ dimensions) concatenated to the MANO input (i.e., $\mathbf y_i  =\mathbf{h}_j=f_{enc}(\boldsymbol\theta_j, \boldsymbol\beta_j,\mathbf c_j)$). Notably, the \emph{contact-oracle} serves as a strong upper bound for conditioning and contact reconstruction quality, which is not available for deployment. In total, $8$ baselines are considered in the ablation studies. For the sampling steps, to balance between the quality convergence and the evaluation time, \emph{HOGraspFlow} samples with $40$ steps, while \emph{HOGraspDiff} requires $100$ steps. The Euler sampler is used for both.

\paragraph{Metrics} Our evaluation metrics involve: (i) earth mover's distance (EMD), which quantifies the mismatch between predicted and ground-truth grasp poses via $SE(3)$ geodesic (lower means better), where we generated $100$ grasps for each data sample for the measurement (following \cite{lim2024equigraspflow}); 
(ii) Contact Accuracy (\emph{CA}$(\%)$), measuring the accuracy of reconstructed contact. In addition, the classification accuracy with respect to the grasp taxonomy (\emph{TA}$(\%)$) serves as a complementary metric. We evaluate each baseline on $26$k validation samples from HOGraspNet at two granularities: (i) overall performance on the full validation split (Tab.~\ref{tab:overall}) in parallel with the EMD distribution for each baseline (Fig.~\ref{fig:emd_dist}); (ii) per grasp type performance (Tab.~\ref{tab:per_type}) across eight representative grasp classes selected to evenly cover the power, intermediate, and precision categories \cite{feix2015grasp}.

\paragraph{Results and analysis}
Tab.~\ref{tab:overall} reports the overall performance across all 8 baseline approaches. In the \emph{HOGraspDiff} branch, the EMD drops from \(1.02\) to \(0.82\) when semantic cues from DINOv2 are integrated. This is further improved by \(0.02\) with the taxonomy-aware codebook. Moreover, A significant increase in the contact accuracy is identified given the semantic feature by over $15\%$ compared to embedding from MANO only input, and has only $1-2\%$ loss compared to the \emph{contact-oracle} model. In contrast, the \emph{HOGraspFlow} branch shows simultaneous gains: the mean drops from \(0.81\) to \(0.69\), and finally reached \(0.67\) with the integrated codebook. Regarding the classification performance, starting from MANO-only, taxonomy accuracy climbs up by over $25\%$ when semantics are embedded in both denoising approaches. In comparison, the \emph{contact-oracle} attains high CA but around $78\%$ in TA, suggesting that per-vertex contact alone is less discriminative of grasp category than visual semantic cues. 

Besides, Fig.~\ref{fig:emd_dist} presents the concrete EMD distributions of each baseline. Across the board, adding RGB semantics produces a clear down-shift of the EMD. The taxonomy-aware codebook yields a consistent additional shift upon the others, represented by higher frequencies in the central area of distributions. \emph{HOGraspFlow} consistently dominates \emph{HOGraspDiff} across baselines, with lower central tendency and tighter spread in terms of interquartile range. Furthermore, Tab.~\ref{tab:per_type} further reports per grasp type results and shows that the trends above hold across the grasp taxonomy. Consistent with Table~\ref{tab:overall}, adding RGB-derived semantic features and CB yields performance comparable to the \emph{contact-oracle} model even under strong occlusions, while preserving low EMD. This justified that contact-only supervision is inadequate for semantic understanding compared with integrating object-aware visual priors. 

In line with the findings from \cite{lim2024equigraspflow}, we observe a significant EMD gap of over \(0.15\) between flow- and score-based models under identical encoding and denoising architectures. Our key insight is that, although both approaches aim to approximate the same data distribution through a time-dependent vector field, they differ fundamentally in the nature of their denoising targets. The SM models learns the score of perturbed marginals under a variance-exploding (VE) perturbation on the Lie algebra. This score is intrinsically heteroskedastic, since its typical norm scales as \(1/\sigma(t)\propto t^{-1/2}\) for both translation and rotation, and its variance increases with \(t\) as \(p_t(\cdot| x_0)\) spreads along the score directions of \(\mathrm{SE}(3)\) space. In contrast, FM directly regresses the instantaneous velocity induced by a chosen bridge between the same marginals, parameterized by decoupled translational and rotational elements (Eq.~\ref{eq:decoupled_flow}) instead of explicit $SE(3)$ geodesics. Hence, the training reduces to unscaled standard regression, where every time step exposes the network to the same target magnitude and direction. This eliminates the scale drift present in score-based convergence and sampling, yielding a well-conditioned optimization on the Lie algebra.

\subsection{Real-world Performance}
\label{subsec:performance2}

\begin{figure}
\hspace{-.3cm}
    \includegraphics[width=1.05\linewidth]{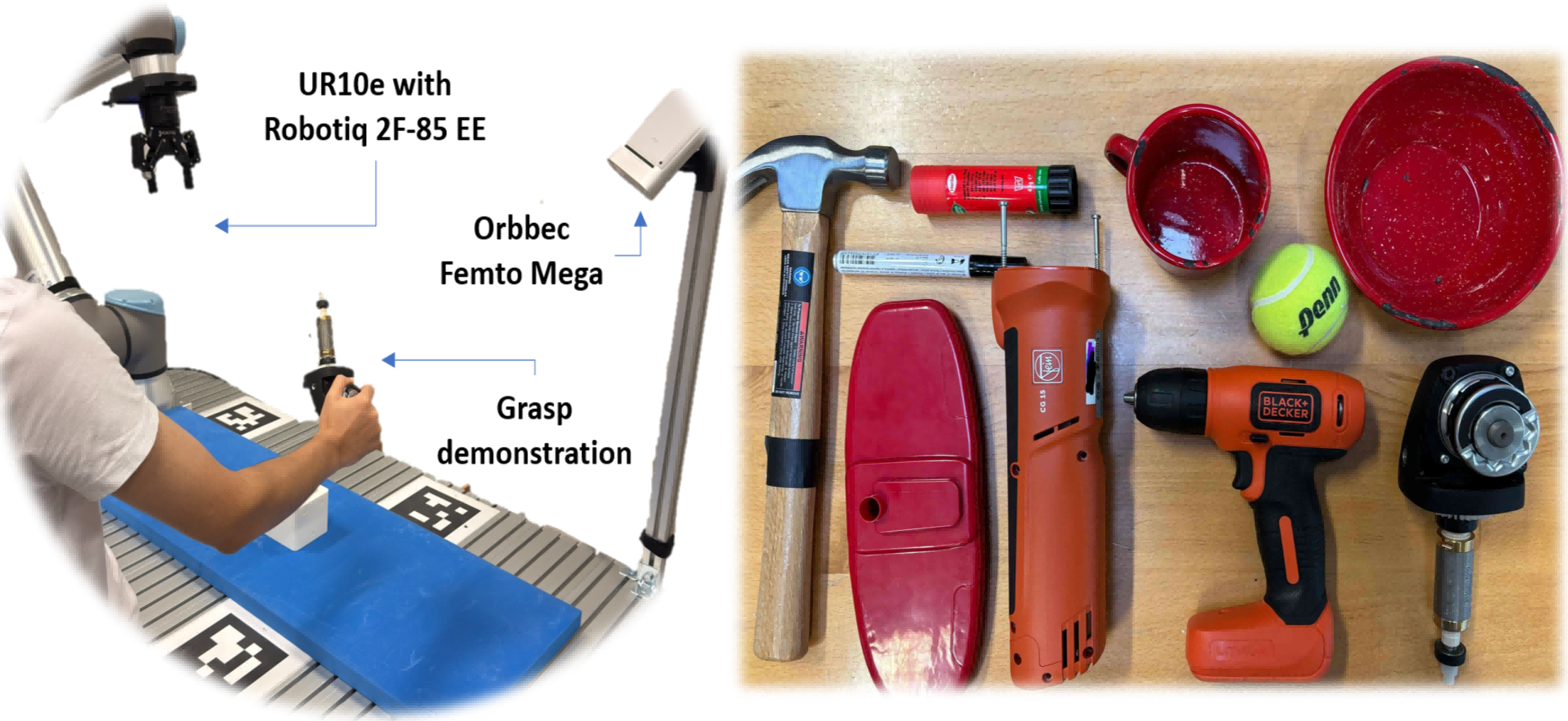}
    \caption{Experiment setup (left) and objects in real-world experiments (right). The hammer, marker, mug and bowl are from the HOGraspNet (\emph{In dist.}), others are OOD (\emph{Out dist.}).}
    \label{fig:objects}
\end{figure}
\paragraph{Setups}
We further conducted real-world experiments to evaluate the practical quality of our generated grasps. The experimental setup consists of a UR10e manipulator equipped with a Robotiq 2F-85 EE. A static Orbbec Femto Mega is used for perception. The object set in the experiments (shown in Fig.~\ref{fig:objects}) consists of a representative set of daily objects, common tools and mechanical parts, which demand a wide range of grasping strategies in practice, spanning from firm power grasps to precise and fine manipulations. Each object was evaluated using $2$ selected grasp types, with $4$ trials per grasp type, chosen to match its functional utility.

\paragraph{Performance}
Tab.~\ref{tab:realworld} summarizes the real-world performance under the baselines of \emph{HOGraspFlow}, where our approach achieved $83.8\%$ ($67/80$) in grasp success rate with $200$ms latency on grasp synthesis, outperforming other aforementioned baselines. We also incorporate: (i) Thumb-index template, adopted from~\cite{wang2025gat, lepert2025phantom}; (ii) GraspNet-1B \cite{fang2020graspnet}, a point cloud-based grasp generation baseline, filtering its candidate grasps to match the observed hand approach direction and contact regions. While GraspNet-1B yields dense and robust grasps on clean, fully visible point clouds, it struggles to balance high grasp confidence with hand-alignment constraints after filtering and is sensitive to point artifacts (e.g., shiny angle grinder motor, the thin body of a pen), reaching $66.2\%$ ($53/80$) overall success.

Fig.~\ref{fig:teaser} further demonstrates qualitative grasp retargeting results, including power, tool-oriented, and fine manipulation grasps. Notably, given reliable \emph{ICP} alignment, our method remains robust due to ignorance of sensory artifacts, whereas geometry-centric approaches often fail due to perception errors. In general, we show the flexibility and generalization of our approach in retargeting diverse human grasps to a PJ EE without accurate geometry or pose estimation on objects. 

Nevertheless, three main failure types are identified: (i) imperfect hand-pose estimation from WiLoR, which can propagate through the entire pipeline; (ii) \emph{ICP} registration errors of hands; (iii) motion planning failures during grasp executions.

\begin{table}[h!]
\setlength{\tabcolsep}{5pt}
\centering
\begin{tabular}{lccc}
\toprule
Baseline  &  \emph{Overall} & \emph{In dist.} & \emph{Out dist.}  \\
\midrule
Thumb-index template~\cite{wang2025gat, lepert2025phantom} & $33/80$ &$13/32$ &$20/48$ \\
GraspNet-1B~\cite{fang2020graspnet}& $53/80$ &$21/32$ &$32/48$ \\ 
\midrule
\emph{HOGraspFlow}, w/o DINOv2  & $55/80$& $18/32$& $37/48$ \\
w/ DINOv2  & $ 63/80$&$21/32$&$\mathbf{42/48}$  \\
w/ DINOv2 CB  & $\mathbf{67/80} $&$\mathbf{26/32}$&$41/48$ \\
\midrule
\bottomrule
\end{tabular}
\caption{Success in real-world grasp retargeting.}
\label{tab:realworld}
\end{table}

\section{CONCLUSIONS}
We proposed \emph{HOGraspFlow}, a vision-based, hand pose-centric retargeting framework that converts a single RGB HOI frame into multi-modal $SE(3)$ parallel jaw grasps. By combining foundational RGB features with a learned contact decoder and a taxonomy-aware codebook, our method injects intent priors that improve distributional fidelity and semantic correctness, with FM consistently outperforming the score-based variant while maintaining high contact and taxonomy accuracy. Real-world experiments confirm robust transfer across diverse objects and grasp types with over $83\%$ success rate, which outperforms the existing template-based proxies and point-based grasp learning approach. 


\bibliographystyle{IEEEtran}
\bibliography{bibfile}

\end{document}